\pdfoutput=1
\documentclass[11pt]{article}

\usepackage[]{latex/acl}

\usepackage{times}
\usepackage{latexsym}
\usepackage{booktabs}
\usepackage{graphicx} 
\usepackage{amsmath} % assumes amsmath package installed
\usepackage{amssymb}  % assumes amsmath package installed
\usepackage{mdframed}
\usepackage{xcolor}
\usepackage{pifont} 
\usepackage{latexsym}
\usepackage{authblk}
\usepackage[T1]{fontenc}
\usepackage[utf8]{inputenc}
\usepackage{fdsymbol}

\usepackage{microtype}
\usepackage{multirow}
\usepackage{multicol}
\usepackage{inconsolata}
\usepackage{enumitem}

\title{LLM-Based Multi-Hop Question Answering with Knowledge Graph Integration in Evolving Environments}
\newcommand{\figref}[1]{Figure~\ref{#1}}

\newcommand{\ie}{\textrm{i.e.}}

\newcommand{\cmark}{\ding{52}}%
\newcommand{\xmark}{\ding{54}}%
\def\reduparrow{\textcolor{green}{\boldsymbol{\uparrow}}}
\def\greendownarrow{\textcolor{red}{\boldsymbol{\downarrow}}}

\def\weblink{\urllink[pre = \bgroup\bf, post = \egroup]}

  {\begin{mdframed}[topline=false, bottomline=false, rightline=false,%
    linewidth=1pt,linecolor=black,%
    leftmargin=1cm,rightmargin=1cm,%
    innerleftmargin=10pt,innerrightmargin=10pt,%
    innertopmargin=0pt,innerbottommargin=0pt,%
    skipabove=\baselineskip,skipbelow=\baselineskip]%
   \fontfamily{ppl}\selectfont% Select Palatino font for the quote
  }%
  {\end{mdframed}}

\newmdenv[
  linecolor=black,
  leftmargin=0,
  rightmargin=0,
  backgroundcolor=gray!10, % Light gray background
  innertopmargin=5pt,
  innerbottommargin=5pt,
  skipabove=\parsep,
  skipbelow=\parsep
]{examplebox}

\author{\bf Ruirui Chen$^1$, Weifeng Jiang$^3$, Chengwei Qin$^3$, Ishaan Singh Rawal$^{1, 2, 4}$,\authorcr \bf Cheston Tan$^{1, 2}$, Dongkyu Choi$^1$, Bo Xiong$^5$, Bo Ai$^{1, 2, 6}$\\
\textsuperscript{1} Institute of High Performance Computing (IHPC) and \textsuperscript{2}Centre for Frontier AI Research,\\
Agency for Science, Technology and Research (A*STAR)\\
\textsuperscript{3}Nanyang Technological University \textsuperscript{4}Texas A\&M University\\
\textsuperscript{5}University of Stuttgart \textsuperscript{6}University of California San Diego
}
\begin{document}
\maketitle
\begin{abstract}
The important challenge of keeping knowledge in Large Language Models (LLMs) up-to-date has led to the development of various methods for incorporating new facts. However, existing methods for such knowledge editing still face difficulties with multi-hop questions that require accurate fact identification and sequential logical reasoning, particularly among numerous fact updates. To tackle these challenges, this paper introduces Graph Memory-based Editing for Large Language Models (GMeLLo), a straightforward and effective method that merges the explicit knowledge representation of Knowledge Graphs (KGs) with the linguistic flexibility of LLMs. Beyond merely leveraging LLMs for question answering, GMeLLo employs these models to convert free-form language into structured queries and fact triples, facilitating seamless interaction with KGs for rapid updates and precise multi-hop reasoning. Our results show that GMeLLo significantly surpasses current state-of-the-art (SOTA) knowledge editing methods in the multi-hop question answering benchmark, MQuAKE, especially in scenarios with extensive knowledge edits.
\end{abstract}

\section{Introduction}

\begin{figure}[t]
    \centering
    \includegraphics[width=\columnwidth]{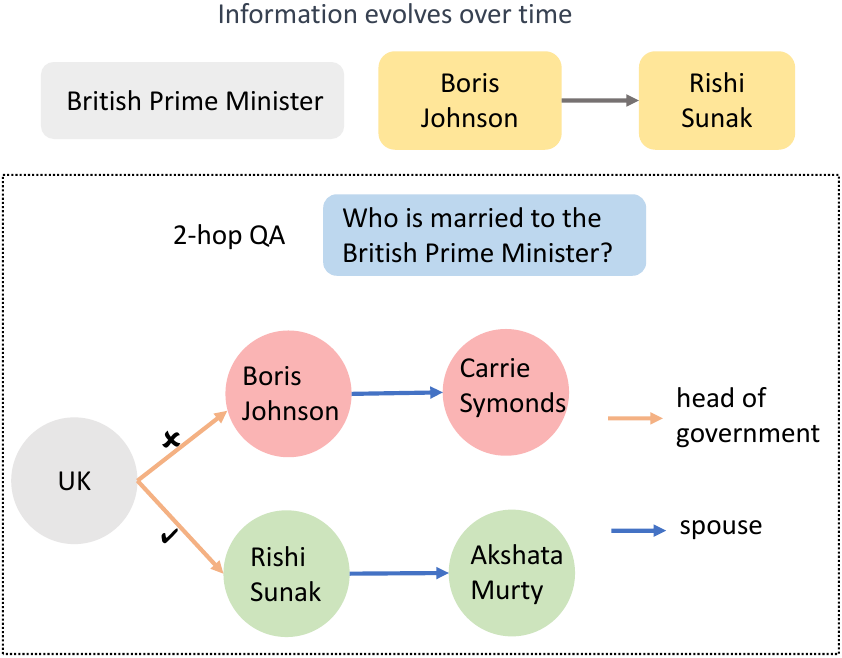}
    \caption{Multi-hop question answering in dynamic domains \cite{zhong2023mquake}. 
    Dynamic nature of information: Changes over time may trigger subsequent modifications. For instance, a transition in the British Prime Minister, such as from Boris Johnson to Rishi Sunak, necessitates corresponding adjustments, like the change in the British Prime Minister's spouse.}
    \label{Multi-hop}
\end{figure}

An important challenge in deploying Large Language Models (LLMs) is keeping their knowledge accurate and up-to-date, without incurring expensive retraining costs~\cite{Sinitsin2020Editable}. Several approaches have been proposed in prior works to address this challenge. Some methods focus on the incremental injection of new facts into language models \cite{52308, de-cao-etal-2021-editing, NEURIPS2022_6f1d43d5, mitchell2022fast}. Alternatively, other methods involve the use of external memory to store new facts \cite{mitchell2022memory, zhong2023mquake}, which does not require updating LLM model weights.

As LLMs operate as black boxes, modifying one fact might inadvertently alter another, making it challenging to guarantee accurate revisions. In this paper, we introduce GMeLLo, an effective approach designed to synergize the strengths of LLMs and Knowledge Graphs (KGs) in addressing the multi-hop question answering task after knowledge editing \cite{zhong2023mquake}. An illustrative example is presented in Figure~\ref{Multi-hop}. Following an information update regarding the British Prime Minister, it becomes evident that the corresponding spouse information should also be modified. 

As depicted in Figure~\ref{GMeLLo}, our GMeLLo method comprises the following key steps:
\begin{itemize}
\item We utilize LLMs to translate edited fact sentences into triples, employing these triples to update the KG and ensure its information remains up to date.
\item Given a question, we utilize LLMs to extract its relation chain, encompassing the primary entity and its connections with other unknown entities. After populating a template, we convert the relation chain into a formal query and use it to search the updated KG.
\item In addition, we retrieve the most pertinent edited facts based on the question and prompt LLMs to generate an answer in accordance with these facts.
\item In instances where the answer provided by the LLM conflicts with that from the KG, we prioritize the answer from the KG as the final response.
\end{itemize}

LLMs, trained on extensive sentence corpora \cite{NEURIPS2020_1457c0d6, rae2022scaling, chowdhery2023palm}, are expected to encapsulate a wide range of commonly used sentence structures. As a result, they are invaluable tools for analyzing sentences and extracting entities and relations. Once the correct relation chain and edited triples are obtained, using a formal query to interrogate the KG in a Knowledge-based Question Answering (KBQA) \cite{10.14778/3055540.3055549} manner ensures precision in the searching process. In cases where KBQA fails, we still have LLMs for question answering (QA) to ensure comprehensive coverage.
GMeLLo outperforms current SOTA methods on two datasets from the MQuAKE benchmark, affirming its effectiveness in multi-hop question answering within an evolving environment.

\section{Related Work}
This work utilizes both KGs and LLMs to address the challenge of multi-hop question answering, with a particular focus on scenarios involving evolving factual knowledge. Therefore, we review existing literature on multi-hop question answering, knowledge editing, and the augmentation of LLMs with knowledge graphs\footnote{Due to space constraints, some of the literature is located in Appendix \ref{dist}.}.

\subsection{Multi-Hop Question Answering}

Multi-hop question answering is more challenging because it requires not only recalling facts but also appropriately aggregating and chaining them. Facts can be sourced from a knowledge graph \citep{lin-etal-2018-multi, 10476130, zhong2023mquake}, tables \citep{yin-etal-2016-neural}, free-form text \citep{yang-etal-2018-hotpotqa, welbl-etal-2018-constructing}, or a heterogeneous combination of these sources \citep{chen-etal-2020-hybridqa, mavi2022survey, lei-etal-2023-s3hqa}. With the development of LLMs, prompt-based methods combined with an optional retrieval module have become a popular approach for handling multi-hop question answering \cite{khattab2022demonstrate, press-etal-2023-measuring, zhong2023mquake}. While most previous works focus on a static information base, our approach targets a dynamic domain, accommodating changes in facts.

\subsection{Knowledge Editing}
As highlighted in \citet{DBLP:journals/corr/abs-2305-13172}, two paradigms exist for editing knowledge: modifying model parameters and preserving model parameters. 

\subsubsection{Parameter-Modification Paradigm}
In the case of modifying model parameters, this can be further categorized into meta-learning or locate-and-edit approaches. Meta-learning methods \cite{de-cao-etal-2021-editing, mitchell2022fast} utilize a hyper network to learn the necessary adjustments for editing LLMs. The locate-then-edit paradigm \cite{dai-etal-2022-knowledge, NEURIPS2022_6f1d43d5, meng2022memit, li2023pmet, gupta2023editing, DBLP:journals/corr/abs-2402-13593} involves initially identifying parameters corresponding to specific knowledge and subsequently modifying them through direct updates to the target parameters.

\subsubsection{Parameter-Preservation Paradigm}
In the case of preserving model parameters, the introduction of additional parameters or external memory becomes necessary. The paradigm of additional parameters \cite{dong-etal-2022-calibrating, hartvigsen2023aging, huang2023transformerpatcher} incorporates extra trainable parameters into the language model. These parameters are trained on a modified knowledge dataset, while the original model parameters remain static. In contrast, memory-based models \cite{mitchell2022memory, zhong2023mquake, gu-etal-2024-pokemqa} explicitly store all edited examples in memory and employ a retriever to extract the relevant edit facts for each new input, guiding the model in generating the updated output.

While previous evaluation paradigms have primarily focused on validating the recall of edited facts, \citet{zhong2023mquake} introduced MQuAKE, a benchmark that includes multi-hop questions involving counterfactual or temporal edits. The two datasets within MQuAKE assess whether methods can accurately answer questions where the response should change due to edited facts. 

\begin{figure*}[t]
    \centering
    \includegraphics[width=2\columnwidth]{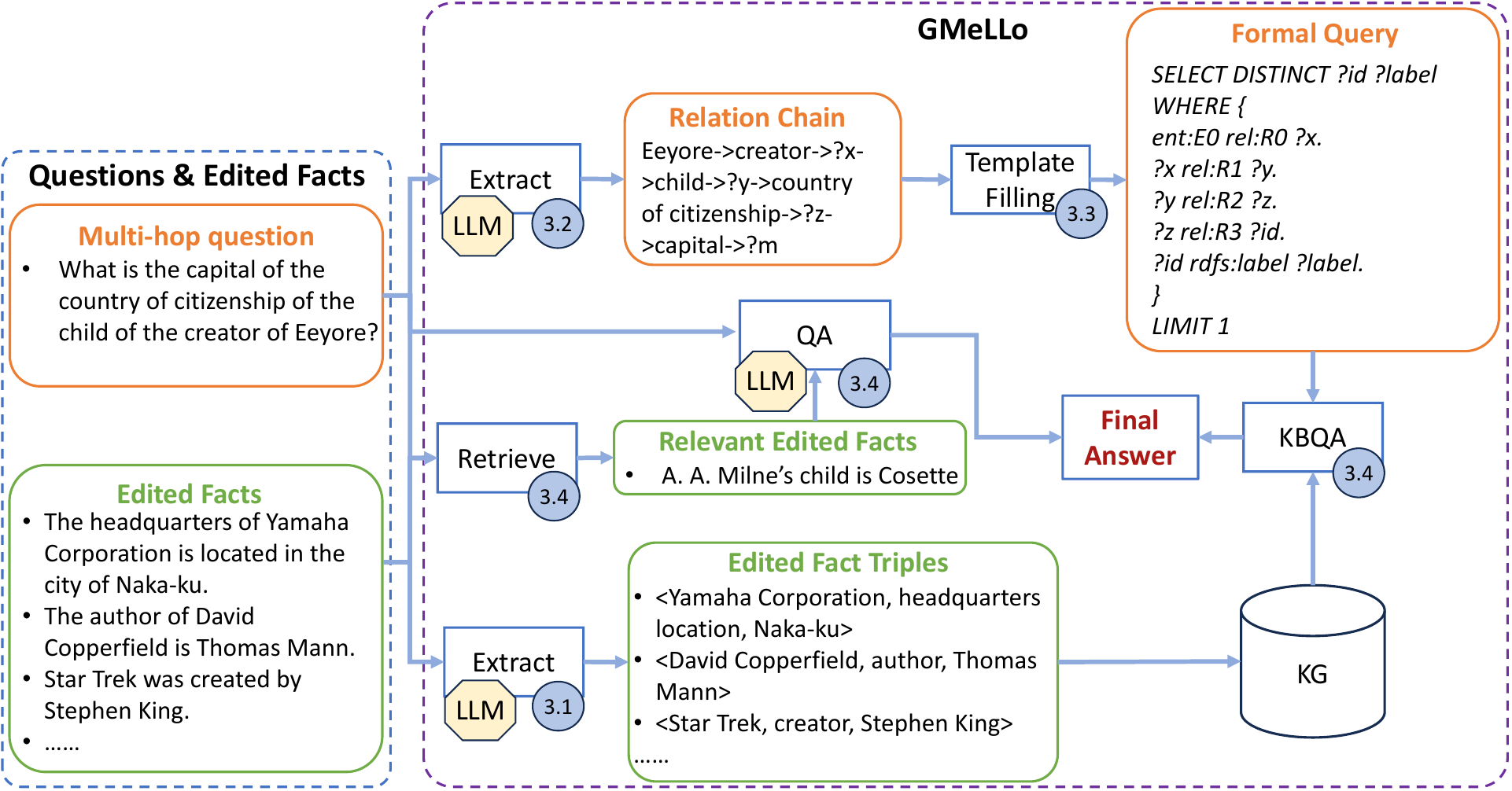}
    \caption{The illustration depicts our proposed method, GMeLLo. We begin by utilizing LLMs to extract entities and relations from edited facts, resulting in a list of edited fact triples. These triples are then used to update a KG.
    Similarly, we employ LLMs to extract relation chains from a given question. By populating this information into a template, we generate a formal query suitable for use in KBQA \cite{Lan-KBQASurvey-2021}.
    Simultaneously, we utilize LLMs for question answering, providing an answer based on the relevant edited facts retrieved. In cases where the LLM's answer contradicts that of the KG, we defer to the KG's answer as the final response. }
    \label{GMeLLo}
\end{figure*}

\section{GMeLLo: Graph Memory-based Editing for Large Language Models}

In this section, we introduce our method GMeLLo for multi-hop question answering with knowledge editing (\figref{GMeLLo}).

\subsection{Extracting Fact Triples from Edited Information Using LLMs} \label{sec:updating-kg}

KGs play a pivotal role in enhancing the capabilities of LLMs by offering external knowledge for improved inference and interpretability, as demonstrated by recent studies \cite{pan2023unifying, rawte2023survey}. Apart from merely storing updated information in an external memory, such as a list of separate sentence statements as seen in conventional approaches \cite{zhong2023mquake}, we utilize the KG to maintain inherent connections and ensure the integration of the latest information.

In our approach, we leverage Wikidata \cite{vrandevcic2014wikidata}, a widely recognized KG, as the foundational knowledge base. When updated facts are received, we utilize LLMs to extract entities from the sentences and determine their relationships (selecting a relation from the predefined list). This process generates edited fact triples, which are then used to update the KG (see Figure~\ref{GMeLLo}). Updating the KG with an edited fact triple involves identifying the connections in the KG based on the subject entity and relation, breaking these connections, and establishing a new connection based on the triple.

We incorporate in-context learning \cite{dong2023survey} to ensure the LLMs have thorough understanding of the task. Furthermore, given the possibility that LLMs may generate relations not present in the predefined relation list \cite{chen2024large}, we use a retrieval model to identify the most similar relation (\ie, the closest relation in the embedding space) from the predefined relation list. The integration of retrieval model makes the triple extraction process more robust.

\subsection{Extracting Relation Chain from Questions Using LLMs} \label{sec:relation-extraction}
As the world evolves rapidly, the training data for LLMs can quickly become outdated. However, since the evolution of linguistic patterns typically progresses at a slower pace, the extensive training data of LLMs should enable them to effectively comprehend most sentence patterns. In this paper, we employ LLMs to extract the relation chain from a sentence, encompassing the mentioned entity in the question and its relations with other unidentified entities. Similar to the fact triple exaction mentioned in Section \ref{sec:updating-kg}, we task LLMs with selecting a relation from a predefined list to mitigate varied representations of the same relation. Take a question sentence from the MQuAKE-CF \cite{zhong2023mquake} dataset as an example,

\begin{examplebox}
    \textbf{Question} \\ 
    \textit{What is the capital of the country of citizenship of the child of the creator of Eeyore?}   \\
    % \vspace{5pt} \\
    \textbf{Relation Chain} \\ 
    \texttt{Eeyore->creator->?x->child->?y\\
    ->country of citizenship\\
    ->?z->capital->?m}
\end{examplebox}

The presented question necessitates a 4-hop reasoning process. With "Eeyore" as the known entity in focus, the journey to the final answer involves identifying its creator "?x", moving on to the creator's child "?y", obtaining the child's country of citizenship "?z", and culminating with the retrieval of the country's capital "?m". All the relations, such as "creator", "child", "country of citizenship", and "capital", are chosen from a predefined list of relations. The relation chain encapsulates all essential information for deriving the answer.

To enable LLMs to extract relation chains and generate outputs in a structured template, we provide several examples of relation chain extraction in the prompt and utilize in-context learning \cite{dong2023survey}, as detailed in Appendix \ref{A3}.

\subsection{Converting a Relation Chain into a Formal Query} \label{sec:query}

Once the relation chain is obtained, the next step involves integrating the known entity and the relations into a formal query template. For a KG represented in RDF\footnote{\url{https://www.w3.org/RDF/}} format, the relation chain elucidated in Section \ref{sec:relation-extraction} can be represented as the following SPARQL\footnote{\url{https://www.w3.org/TR/sparql11-query/}} query, 
% underscoring the seamless integration of the obtained information into a structured query framework.

\begin{quote}
\begin{verbatim}
PREFIX ent: <http://www.kg/entity/>
PREFIX rel: <http://www.kg/relation/>
SELECT DISTINCT ?id ?label WHERE {
   ent:E0 rel:R0 ?x.
   ?x rel:R1 ?y.
   ?y rel:R2 ?z.
   ?z rel:R3 ?id.
   ?id rdfs:label ?label.
}
LIMIT 1
\end{verbatim}
\end{quote}

In this context, "ent" and "rel" serve as prefixes for entity and relation, respectively. The identifier "E0" uniquely represents "Eeyore" within the KG, while the identifiers for "creator," "child," "country of citizenship," and "capital" are denoted as "R0", "R1", "R2", and "R3", respectively. After identifying the entity "?id", we retrieve its string label "?label" as the final answer.

\subsection{Integrating LLM-based QA and KBQA}
This subsection outlines the integration of the proposed KBQA module with the LLM-based QA module within the GMeLLo framework.

\textbf{LLM-based question answering.} 
When a question arises, we retrieve the top-$x$
relevant facts using the pre-trained Contriever \cite{izacard2022unsupervised} model from a list of edited fact sentences. We then prompt the LLMs to generate answers based on the question and these pertinent facts. Compared to the "split-answer-check" pipeline in MeLLo \cite{zhong2023mquake}, this LLM-based QA method is expected to be simpler and yield more accurate results when the facts are provided accurately. 

However, addressing multi-hop questions, especially those where the edited facts pertain to intermediary hops, presents a challenge in accurately retrieving the relevant information and performing correct multi-hop question answering. This challenge is particularly pronounced when dealing with a large volume of edited facts. For instance, accurately identifying the relevant fact given the question in Figure~\ref{GMeLLo} and producing the correct final answer is difficult. 

\textbf{KBQA.}
To address the challenges of LLM-based question answering, we integrate responses from KBQA to refine the outputs from the LLMs, as detailed in the previous section. When the relation chain and fact triples are accurately derived, the KBQA system provides the correct answer. However, if the relation chain is incorrectly extracted, the search path in the KG may become invalid, leading the KBQA system to yield no output. In such instances, we accept the response from the LLMs as the final answer.

\section{Experiment}
In this section, we will present the results from our experiments to demonstrate the effectiveness of employing our GMeLLo methodology.

\subsection{Experiment Setup}

\subsubsection{Dataset}
Our experiment focuses on the multi-hop question-answering benchmark, MQuAKE \cite{zhong2023mquake}, which comprises two datasets: MQuAKE-CF\footnote{Following \citet{zhong2023mquake}, our experiments on MQuAKE-CF are carried out on a randomly sampled subset of the complete dataset, comprising 3000 instances in total(1000 instances for each of {2, 3, 4}-hop questions).}, designed for counterfactual edits, and MQuAKE-T, specifically tailored for updates in temporal knowledge.

%Table~\ref{data_statistics} provides a summary of the statistics for the MQuAKE-CF and MQuAKE-T datasets. 

The MQuAKE-CF dataset comprises 3,000 N-hop questions (\(\textrm{N} \in \{2,3,4\}\)), each linked to one or more edits. This dataset functions as a diagnostic tool for examining the effectiveness of knowledge editing methods in handling counterfactual edits. The MQuAKE-T dataset consists of 1,868 instances, each associated with a real-world fact change. Its purpose is to evaluate the efficacy of knowledge editing methods in updating obsolete information with contemporary, factual data. A table of statistics is available in Appendix \ref{A1}. 

\subsubsection{Evaluation Settings}
To evaluate our models, we adhere to the testing settings outlined by \citet{zhong2023mquake}. Specifically, instances are batched in groups of size $k$, with $k \in {1, 100, 1000, 3000}$ for MQuAKE-CF, and $k \in {1, 100, 500, 1868}$ for MQuAKE-T. For example, in the MQuAKE-CF dataset, when $k=100$, the $3000$ instances are split into $30$ groups, and we report the average performance as the final result. 

For each test instance, the dataset includes three multi-hop questions that convey the same meaning. In alignment with \citet{zhong2023mquake}, if the model correctly answers any one of these questions, we consider the instance to be accurately resolved.

\subsubsection{Baselines}
To demonstrate the effectiveness of our approach, we conduct comparisons with the following SOTA knowledge editing methods.
\begin{itemize}
    \item MEND \cite{mitchell2022fast}. It trains a hyper-network to generate weight updates by transforming raw fine-tuning gradients based on an edited fact.
    \item MEMIT \cite{meng2022memit}. It updates feed-forward networks across various layers to incorporate all relevant facts.
    \item MeLLo \cite{zhong2023mquake}. It employs a memory-based approach for multi-hop question answering, storing all updated facts in an external memory. 
    \item PokeMQA \cite{gu-etal-2024-pokemqa}. It also uses a memory-based approach, which decouples question decomposition from knowledge editing to reduce the burden on LLMs. Additionally, it introduces auxiliary knowledge prompts to assist with question decomposition.
\end{itemize}

Given the substantial costs associated with training, deploying, and maintaining larger LLMs \cite{li2023textbooks}, and the challenges of scaling up knowledge editing methods that require model parameter modifications, this paper primarily focuses on smaller LLMs, specifically GPT-J (6B) \cite{gpt-j} and Vicuna (7B) \cite{vicuna2023}. However, to showcase GMeLLo's effectiveness with larger LLMs in practical scenarios, we also report the performance of both MeLLo and GMeLLo on the MQuAKE-CF dataset when $k=3000$.

% Table generated by Excel2LaTeX from sheet 'Main_performance'
\begin{table*}[t]
  \centering
    \begin{tabular}{llrrrrrrrr}
    \toprule
    \multicolumn{1}{c}{\multirow{2}[0]{*}{Base Model}} & \multicolumn{1}{c}{\multirow{2}[0]{*}{Method}} & \multicolumn{4}{c}{MQuAKE-CF} & \multicolumn{4}{c}{MQuAKE-T} \\
          &       & \multicolumn{1}{l}{k=1} & \multicolumn{1}{l}{k=100} & \multicolumn{1}{l}{k=1000} & \multicolumn{1}{l}{k=3000} & \multicolumn{1}{l}{k=1} & \multicolumn{1}{l}{k=100} & \multicolumn{1}{l}{k=500} & \multicolumn{1}{l}{k=1868} \\
    \midrule
    \multirow{4}{*}{GPT-J-6B} & MEMIT & 12.3  & 9.8   & 8.1   & 1.8   & 4.8   & 1.0     & 0.2   & 0.0 \\
     & MEND  & 11.5  & 9.1   & 4.3   & 3.5   & 38.2  & 17.4  & 12.7  & 4.6 \\
     & MeLLo & 20.3  & 12.5  & 10.4  & 9.8   & 85.9  & 45.7  & 33.8  & 30.7 \\
     & \textbf{GMeLLo} & \textbf{76.3}    & \textbf{53.4}  & \textbf{49.5} & \textbf{49.0}  &  \textbf{86.9}   &   \textbf{82.1}    &   \textbf{81.5}    &  \textbf{81.5}\\
    \midrule
    \multirow{2}{*}{Vicuna-7B} & MeLLo & 20.3  & 11.9  & 11.0    & 10.2  & 84.4  & 56.3  & 52.6  & 51.3 \\
    &PokeMQA & 45.8  & 38.8  & -    & 31.6  & 74.6  & -  & -  & 73.1 \\
     & \textbf{GMeLLo} &  \textbf{71.3}    &  \textbf{46.5}   &    \textbf{42.5}  & \textbf{41.9}   &   \textbf{97.1}    &   \textbf{86.3}   & \textbf{85.4}  & \textbf{85.1} \\
    \bottomrule
    \end{tabular}
    \caption{Performance comparison of GMeLLo and other approaches on the MQuAKE-CF and MQuAKE-T datasets using GPT-J-6B or Vicuna-7B as the base language models. Adhering to the methodology outlined by \citet{zhong2023mquake}, instances are grouped into batches of size $k$. For the MQuAKE-CF dataset, $k$ varies from 1 to 3000, and for the MQuAKE-T dataset, it ranges from 1 to 1868. For example, in the MQuAKE-CF dataset, when $k=100$, the 3000 instances are organized into 30 groups, and the average performance reported as the final result. The metric used is accuracy.}
  \label{main}%
\end{table*}%

\subsubsection{Knowledge Graph Setting}

Considering Wikidata's community-driven nature, guaranteeing a dynamic and comprehensive dataset across a spectrum of knowledge domains, we use Wikidata \cite{vrandevcic2014wikidata} as the foundational KG for this experiment. To align the relations in the question and fact sentences with those in WikiData \cite{vrandevcic2014wikidata}, we take the following steps:
\begin{itemize}
\item First, we select the first 500 item properties\footnote{\url{https://www.wikidata.org/w/index.php?title=Special:ListProperties/wikibase-item\&limit=500\&offset=0}} from WikiData as the base relations. Items represent either concrete or abstract entities, such as a person \cite{10.1145/3306446.3340822}.
\item Next, we employ GPT-3.5-Turbo\footnote{\url{https://platform.openai.com/docs/models/gpt-3-5-turbo}} to examine each multi-hop question in the test samples and determine whether it contains any of the base relations or not.
\item Afterward, we rank the frequencies of each relation and choose the top 50 relations as candidates for use in relation chain extraction and edited fact triple extraction.
\end{itemize}

To stay updated with the latest information on WikiData, we utilize the WikiData API service\footnote{\url{https://www.wikidata.org/w/api.php}} and the WikiData Query Service\footnote{\url{https://query.wikidata.org/sparql}}. The correctness of our KBQA result hinges on the accurate extraction of both edited fact triples and relation chains.
If the relation chain is found to be incorrect, we conduct an online search on WikiData to determine if the relation chain leads to an entity that could potentially yield an incorrect answer for the specific question, which takes about 1 second. 

\subsubsection{Strategies for Managing Unforeseen Relationships}
As previously noted, since LLMs may produce relations that are similar in meaning but not identical, we employ the pretrained Contriever model \cite{izacard2022unsupervised} to retrieve the most similar relation (i.e., the closest relation in the embedding space) from the base list of relations. This replacement is performed when undefined relations are encountered during both edited fact triple extraction and relation chain extraction.

\subsection{Main Results} \label{sec:benchmark}

As shown in Table~\ref{main}\footnote{We use the baseline performance reported in \citet{zhong2023mquake} and \citet{gu-etal-2024-pokemqa}. Since the experiment settings in \citet{gu-etal-2024-pokemqa} differ from those in \citet{zhong2023mquake}, we only include the results from \citet{gu-etal-2024-pokemqa} under the same settings. A dash ('-') indicates that performance was not reported for that setting.}, our GMeLLo significantly outperforms all existing methods on the both the MQuAKE-CF dataset and the MQuAKE-T dataset \cite{zhong2023mquake}, particularly when handling a large number of edits.
\begin{table*}[t]
  \centering
    \begin{tabular}{llcccccccc}
    \toprule
    \multicolumn{1}{c}{\multirow{2}[0]{*}{Base Model}} & \multicolumn{1}{c}{\multirow{2}[0]{*}{Method}} & \multicolumn{4}{c}{MQuAKE-CF} & \multicolumn{4}{c}{MQuAKE-T} \\
    \cmidrule(lr){3-6} \cmidrule(lr){7-10} 
          &       & \multicolumn{1}{l}{k=1} & \multicolumn{1}{l}{100} & \multicolumn{1}{l}{1000} & \multicolumn{1}{l}{3000} & \multicolumn{1}{l}{k=1} & \multicolumn{1}{l}{100} & \multicolumn{1}{l}{500} & \multicolumn{1}{l}{1868} \\
    \midrule
    \multirow{3}{*}{GPT-J-6B} & QA  & 71.0 & 24.2 & 14.3 & 12.2 & 32.3 & 18.0 & 15.7 & 15.5 \\
      & KBQA  & 43.3 & 43.3 & 43.3 & 43.3 & 80.2 & 80.2 & 80.2 & 80.2 \\
      & GMeLLo & \textbf{76.3} & \textbf{53.4} & \textbf{49.5} & \textbf{49.0} & \textbf{86.9} & \textbf{82.1} & \textbf{81.5} & \textbf{81.5} \\
    \midrule
    \multirow{3}{*}{Vicuna-7B} & QA  & \textbf{72.6} & 27.0 & 16.5 & 13.5 & 96.9 & 63.0 & 59.2 & 58.2 \\
      & KBQA  & 35.9 & 35.9 & 35.9 & 35.9 & 73.6 & 73.6 & 73.6 & 73.6 \\
      & GMeLLo & 71.3 & \textbf{46.5} & \textbf{42.5} & \textbf{41.9} & \textbf{97.1} & \textbf{86.3} & \textbf{85.4} & \textbf{85.1} \\
    \bottomrule
    \end{tabular}%
  \caption{Ablation study of GMeLLo. QA involves directly using LLM for answering the multi-hop questions. KBQA involves using LLM to transform edited fact sentences into triples, update WikiData, convert question sentences into relation chains, and generate formal KG queries for question answering. GMeLLo combines these methods by using KBQA to correct answers from LLM-based QA.} 
  \label{ablation}%
\end{table*}%

The performance degradation in MeLLo is primarily due to its challenges in identifying relevant facts as the number of edits increases. When k=1, the model utilizes only the facts directly related to the input question for context. However, as k increases, the model faces the challenge of discerning relevant facts from a broader memory. Our proposed GMeLLo model mitigates this by employing an explicit symbolic graph representation, which enhances the system's ability to update and retrieve relevant facts effectively. This feature significantly boosts the scalability of GMeLLo, making it well-suited for real-world question answering applications that require managing large volumes of rapidly changing information.

To further validate our findings with more capable base models, we evaluated MeLLo and GMeLLo using two larger models, GPT-3.5-Turbo-Instruct and GPT-3.5-Turbo, on the MQuAKE-CF dataset with k=3000\footnote{The model \texttt{text-davinci-003} used in \citet{zhong2023mquake} was deprecated on January 4, 2024.}. The accuracy rates achieved by MeLLo and GMeLLo with GPT-3.5-Turbo-Instruct were 30.7\% and 51.4\%, respectively. While GMeLLo achieved an accuracy of 66.4\% with GPT-3.5-Turbo, the same model consistently returned errors when tested with MeLLo, suggesting that the prompts may require modification for compatibility with chat completion models. These results indicate that GMeLLo performs well even when scaled to larger LLMs.

\subsection{Ablation Study}
To gain a comprehensive understanding of the performance of various components, i.e., LLM-based QA and KBQA, we conduct an experiment to illustrate the impact of LLM-based QA and KBQA as the number of edits increases. 

As demonstrated in Table~\ref{ablation}, the performance of KBQA remains consistent because all edited facts are converted to triples and all relation chains are extracted from the test questions, regardless of the value of "k". Correctly answering a multi-hop question in KBQA requires both accurate extraction of fact triples and the relation chain. However, as the parameter "k" increases, more edited facts are stored in the external memory. Consequently, selecting the relevant edits to accurately answering the questions becomes increasingly challenging for LLM-based QA.

When k=1 and all relevant facts are provided to the LLMs for question answering, the LLM-based QA proves to be quite effective. However, a more realistic scenario involves multiple edits occurring simultaneously, where each question is asked separately (i.e., k>1). The performance showcased in Table \ref{ablation} demonstrates the effectiveness of our GMeLLo, highlighting that KBQA serves as a valuable enhancement to LLM-based QA within evolving environments.

\subsubsection{Further Analysis}
To evaluate the impact of KBQA on LLM-based QA within the GMeLLo framework, we conducted an analysis comparing the responses from LLMs to those from the KG. We consider the KG's response as the final answer. Therefore, comparing to only using LLM-based QA, if the answer from LLMs is correct but the answer from the KG is incorrect, this leads to a decline in performance. Conversely, if the answer from LLMs is incorrect but the answer from the KG is correct, performance improves. If the KBQA provides no response, performance remains unchanged. As illustrated in Table \ref{correction-ratio}, when there are discrepancies between KBQA and LLM-based QA responses, the likelihood of KBQA providing the correct answer increases as the parameter "k" increases.

\begin{table*}[t]
  \centering
    \begin{tabular}{lccccccccccc}
    \toprule
    \multicolumn{1}{c}{\multirow{2}[0]{*}{Base Model}} & \multicolumn{3}{c}{\multirow{1}[0]{*}{Scenario}} & \multicolumn{4}{c}{MQuAKE-CF} & \multicolumn{4}{c}{MQuAKE-T} \\
    \cmidrule(lr){2-4} \cmidrule(lr){5-8} \cmidrule(lr){9-12} 
          & LLM & KG & Performance & \multicolumn{1}{l}{k=1} & \multicolumn{1}{l}{100} & \multicolumn{1}{l}{1000} & \multicolumn{1}{l}{3000} & \multicolumn{1}{l}{k=1} & \multicolumn{1}{l}{100} & \multicolumn{1}{l}{500} & \multicolumn{1}{l}{1868} \\
    \midrule
    \multirow{4}{*}{GPT-J-6B} 
      & \xmark{} & \cmark{} & $\reduparrow{}$ & 8.1 & 22.9 &  24.9 & 25.0 & 44.0 & 47.2 & 47.9 & 48.0   \\
      
      & \cmark{} & \xmark{} & $\greendownarrow{}$ & 12.5 & 2.4 & 1.2 & 0.7 & 0.7 & 0.4 & 0.3 & 0.3 \\
      & \cmark{} & $\bigcirc{}$ & - & 34.2 & 7.0 & 4.0 & 3.7 & 7.1 & 2.8 & 2.4 & 2.3 \\
    \midrule
    \multirow{4}{*}{Vicuna-7B} 
      & \xmark{} & \cmark{} & $\reduparrow{}$ & 7.7 & 17.8 & 19.6 & 20.0 & 4.2 & 19.7 & 21.4 & 21.7 \\
     
      & \cmark{} & \xmark{} & $\greendownarrow{}$ & 21.8 & 3.9 & 2.0 & 1.2 & 7.2 & 4.2 & 4.0 & 3.9 \\
      & \cmark{} & $\bigcirc{}$ & - & 32.7 & 7.4 & 4.0 & 3.4 & 35.7 & 19.8 & 18.1 & 17.7 \\
    \bottomrule
    \end{tabular}%
    \caption{Further analysis for scenarios where the answers from LLM and KG contradict each other. The values are expressed as percentages. It is important to note that the total number of test questions is three times the number of test instances. For instance, in MQuAKE-CF, each test instance comprises three distinct questions with the same meaning, totaling 9,000 test questions. Symbols used: $\reduparrow{}$ indicates improved performance, $\greendownarrow{}$ indicates reduced performance, and $\bigcirc{}$ denotes no response from KBQA, resulting in no impact on the final output (-).} 
  \label{correction-ratio}%
\end{table*}%

\subsection{Qualitative Analysis}
Table~\ref{ablation} illustrates that Vicuna exhibits superior performance in directly handling the QA task, particularly when provided with the exact edited facts. Conversely, GPT-J excels in sentence analysis tasks, showcasing its high performance in the KBQA task.
\subsubsection{Inferior Performance of GPT-J in QA}
Table \ref{ablation} shows that the performance of GPT-J and Vicuna in conducting QA tasks is comparable on the MQuAKE-CF dataset when k=1. However, GPT-J exhibits notably lower performance on the MQuAKE-T dataset. Further analysis revealed that GPT-J struggles in answering questions with only an edited fact pertaining to its intermediary information, such as:

\begin{examplebox}
    \textbf{Sample from MQuAKE-CF} \\ 
    \textbf{Facts: }\textit{Midfielder is associated with the sport of Gaelic football}   \\
    \textbf{Question: }\textit{What is the capital of the country where the sport associated with Kieron Dyer's specialty was first played?}   \\ 
    \textbf{Predicted Answer: }\textit{Bondi Junction}   \\ 
    \textbf{Answer: }\textit{Dublin} 
\end{examplebox}

\begin{examplebox}
    \textbf{Sample from MQuAKE-T} \\ 
    \textbf{Facts: }\textit{The name of the current head of the Philippines government is Bongbong Marcos}   \\
    \textbf{Question: }\textit{Who is the head of government of the country that Joey de Leon is a citizen of?}   \\ 
    \textbf{Predicted Answer: }\textit{Benigno Aquino III}   \\ 
    \textbf{Answer: }\textit{Bongbong Marcos} 
\end{examplebox}

However, it can achieve the correct answer in KBQA because it accurately extracts the fact triple and relation chain of the question. Given that all test samples in MQuAKE-T contain only one edited fact, while approximately 63.6\% of test samples in MQuAKE-CF consist of more than two edited facts, GPT-J is able to connect most of the information together. Therefore, it achieves better performance in the MQuAKE-CF dataset.
\subsubsection{Inferior Performance of Vicuna in KBQA}
Compared to GPT-J, Vicuna performs less effectively in the KBQA task. Aside from misunderstandings, the main reasons are as follows:
\begin{itemize}
\item It often makes errors in the sequence. For example, given the fact "The author of Misery is Richard Dawkins", its output fact triple is "Richard Dawkins->author->Misery". However, the correct sequence is "Misery->author->Richard Dawkins".
\item It frequently makes errors in selecting a relation from the list. For example, it often outputs a relation chain as "Mike->citizenship->country->head of state", instead of "Mike->country of citizenship->head of state".
\end{itemize}
It is important to note that even if the relation chain is incorrect, the KBQA system may still provide the correct answer because of some loops in WikiData, such as the country of the USA is the USA.

Although Vicuna is not as effective overall, we still find that in some cases it can correctly extract relations, but cannot provide the correct answer directly. An example is given as follows:

\begin{examplebox}
    \textbf{Sample from MQuAKE-CF} \\ 
    \textbf{Facts: }\textit{Point guard is associated with the sport of cricket}   \\
    \textbf{Question: }\textit{What is the capital of the country from which Erik Spoelstra's sport comes?}   \\ 
    \textbf{Predicted Answer: }\textit{Miami}   \\ 
    \textbf{Answer: }\textit{London} 
\end{examplebox}
\subsection{Further Discussion}
KG offers a clearer representation of multi-hop information and its updates. In GMeLLo, we harness the strengths of both KBQA and LLM-based QA, benefiting from KBQA's high precision and LLM-based QA's extensive coverage. Our experiments reveal that GPT-J excels in extracting relation chains and fact triples, whereas Vicuna demonstrates superior performance in LLM-based QA. Given that KBQA and LLM-based QA operate as separate modules in GMeLLo, we can optimize their use by employing different LLMs in each module, maximizing their effectiveness in practical applications.
\section{Conclusion}
In this paper, we present GMeLLo, a method designed for multi-hop question answering in dynamic environments. In addition to leveraging LLMs for question answering, we also leverage the capabilities of LLMs to extract the triples from edited fact sentences to update a KG, and use the capabilities of LLMs to analyze question sentences and generate a relation chain, and finally get the formal query by filling in a formal query template. Finally, we combine KBQA and LLM-based QA to bolster the multi-hop question answering capability within a dynamic environment. This approach capitalizes on the strengths of both LLMs and KGs. By utilizing LLMs for analyzing question sentences and QA to ensure the coverage, and KBQA to provide accurate results, we achieve a synergy between these two methodologies.

\section*{Limitations}
Despite the promising results, it is important to acknowledge that this investigation is still in its early stages. Although our performance significantly surpasses baseline approaches in multi-hop questions in dynamic domains, particularly for large knowledge bases and complex questions, there is still room for further improvement. Our future research includes 
\begin{itemize}
    \item Leveraging more sophisticated prompting techniques, such as Chain of Thought (CoT) \cite{NEURIPS2022_9d560961}, to enable more accurate multi-hop reasoning.
    \item Refining the predefined relation list to enhance its accuracy.
    \item Enhancing the KG to support more complex question answering, such as inquiries involving historical information.
\end{itemize}
We believe these improvements can further enhance the performance and scalability of the system, enabling it to handle more complex and diverse real-world applications.

\section*{Acknowledgments}

We thank all reviewers for providing valuable feedback. This work was partially supported by A*STAR CRF funding awarded to Cheston Tan. Bo Xiong is funded by the Deutsche Forschungsgemeinschaft (DFG, German Research Foundation) – SFB-1574 – 471687386.

\bibliography{acl_latex_final}

\begin{thebibliography}{46}
\expandafter\ifx\csname natexlab\endcsname\relax\def\natexlab#1{#1}\fi

\bibitem[{Baek et~al.(2023)Baek, Aji, and Saffari}]{baek-etal-2023-knowledge-augmented}
Jinheon Baek, Alham~Fikri Aji, and Amir Saffari. 2023.
\newblock \href {https://doi.org/10.18653/v1/2023.matching-1.7} {Knowledge-augmented language model prompting for zero-shot knowledge graph question answering}.
\newblock In \emph{Proceedings of the First Workshop on Matching From Unstructured and Structured Data (MATCHING 2023)}, pages 70--98, Toronto, ON, Canada. Association for Computational Linguistics.

\bibitem[{Brown et~al.(2020)Brown, Mann, Ryder, Subbiah, Kaplan, Dhariwal, Neelakantan, Shyam, Sastry, Askell, Agarwal, Herbert-Voss, Krueger, Henighan, Child, Ramesh, Ziegler, Wu, Winter, Hesse, Chen, Sigler, Litwin, Gray, Chess, Clark, Berner, McCandlish, Radford, Sutskever, and Amodei}]{NEURIPS2020_1457c0d6}
Tom Brown, Benjamin Mann, Nick Ryder, Melanie Subbiah, Jared~D Kaplan, Prafulla Dhariwal, Arvind Neelakantan, Pranav Shyam, Girish Sastry, Amanda Askell, Sandhini Agarwal, Ariel Herbert-Voss, Gretchen Krueger, Tom Henighan, Rewon Child, Aditya Ramesh, Daniel Ziegler, Jeffrey Wu, Clemens Winter, Chris Hesse, Mark Chen, Eric Sigler, Mateusz Litwin, Scott Gray, Benjamin Chess, Jack Clark, Christopher Berner, Sam McCandlish, Alec Radford, Ilya Sutskever, and Dario Amodei. 2020.
\newblock \href {https://proceedings.neurips.cc/paper_files/paper/2020/file/1457c0d6bfcb4967418bfb8ac142f64a-Paper.pdf} {Language models are few-shot learners}.
\newblock In \emph{Advances in Neural Information Processing Systems}, volume~33, pages 1877--1901. Curran Associates, Inc.

\bibitem[{Chen et~al.(2024)Chen, Qin, Jiang, and Choi}]{chen2024large}
Ruirui Chen, Chengwei Qin, Weifeng Jiang, and Dongkyu Choi. 2024.
\newblock Is a large language model a good annotator for event extraction?
\newblock In \emph{Proceedings of the AAAI Conference on Artificial Intelligence}, volume~38, pages 17772--17780.

\bibitem[{Chen et~al.(2020)Chen, Zha, Chen, Xiong, Wang, and Wang}]{chen-etal-2020-hybridqa}
Wenhu Chen, Hanwen Zha, Zhiyu Chen, Wenhan Xiong, Hong Wang, and William~Yang Wang. 2020.
\newblock \href {https://doi.org/10.18653/v1/2020.findings-emnlp.91} {{H}ybrid{QA}: A dataset of multi-hop question answering over tabular and textual data}.
\newblock In \emph{Findings of the Association for Computational Linguistics: EMNLP 2020}, pages 1026--1036, Online. Association for Computational Linguistics.

\bibitem[{Cheng et~al.(2023)Cheng, Niu, Mo, and Chen}]{10476130}
Zhen Cheng, Jianwei Niu, Shasha Mo, and Jia Chen. 2023.
\newblock \href {https://doi.org/10.1109/ICPADS60453.2023.00166} {Genboost: Generative modeling and boosted learning for multi-hop question answering over incomplete knowledge graphs}.
\newblock In \emph{2023 IEEE 29th International Conference on Parallel and Distributed Systems (ICPADS)}, pages 1131--1138.

\bibitem[{Chiang et~al.(2023)Chiang, Li, Lin, Sheng, Wu, Zhang, Zheng, Zhuang, Zhuang, Gonzalez, Stoica, and Xing}]{vicuna2023}
Wei-Lin Chiang, Zhuohan Li, Zi~Lin, Ying Sheng, Zhanghao Wu, Hao Zhang, Lianmin Zheng, Siyuan Zhuang, Yonghao Zhuang, Joseph~E. Gonzalez, Ion Stoica, and Eric~P. Xing. 2023.
\newblock \href {https://lmsys.org/blog/2023-03-30-vicuna/} {Vicuna: An open-source chatbot impressing gpt-4 with 90\%* chatgpt quality}.

\bibitem[{Chowdhery et~al.(2023)Chowdhery, Narang, Devlin, Bosma, Mishra, Roberts, Barham, Chung, Sutton, Gehrmann et~al.}]{chowdhery2023palm}
Aakanksha Chowdhery, Sharan Narang, Jacob Devlin, Maarten Bosma, Gaurav Mishra, Adam Roberts, Paul Barham, Hyung~Won Chung, Charles Sutton, Sebastian Gehrmann, et~al. 2023.
\newblock Palm: Scaling language modeling with pathways.
\newblock \emph{Journal of Machine Learning Research}, 24(240):1--113.

\bibitem[{Cui et~al.(2017)Cui, Xiao, Wang, Song, Hwang, and Wang}]{10.14778/3055540.3055549}
Wanyun Cui, Yanghua Xiao, Haixun Wang, Yangqiu Song, Seung-won Hwang, and Wei Wang. 2017.
\newblock \href {https://doi.org/10.14778/3055540.3055549} {Kbqa: learning question answering over qa corpora and knowledge bases}.
\newblock \emph{Proc. VLDB Endow.}, 10(5):565–576.

\bibitem[{Dai et~al.(2022)Dai, Dong, Hao, Sui, Chang, and Wei}]{dai-etal-2022-knowledge}
Damai Dai, Li~Dong, Yaru Hao, Zhifang Sui, Baobao Chang, and Furu Wei. 2022.
\newblock \href {https://doi.org/10.18653/v1/2022.acl-long.581} {Knowledge neurons in pretrained transformers}.
\newblock In \emph{Proceedings of the 60th Annual Meeting of the Association for Computational Linguistics (Volume 1: Long Papers)}, pages 8493--8502, Dublin, Ireland. Association for Computational Linguistics.

\bibitem[{De~Cao et~al.(2021)De~Cao, Aziz, and Titov}]{de-cao-etal-2021-editing}
Nicola De~Cao, Wilker Aziz, and Ivan Titov. 2021.
\newblock \href {https://doi.org/10.18653/v1/2021.emnlp-main.522} {Editing factual knowledge in language models}.
\newblock In \emph{Proceedings of the 2021 Conference on Empirical Methods in Natural Language Processing}, pages 6491--6506, Online and Punta Cana, Dominican Republic. Association for Computational Linguistics.

\bibitem[{Dong et~al.(2022)Dong, Dai, Song, Xu, Sui, and Li}]{dong-etal-2022-calibrating}
Qingxiu Dong, Damai Dai, Yifan Song, Jingjing Xu, Zhifang Sui, and Lei Li. 2022.
\newblock \href {https://doi.org/10.18653/v1/2022.findings-emnlp.438} {Calibrating factual knowledge in pretrained language models}.
\newblock In \emph{Findings of the Association for Computational Linguistics: EMNLP 2022}, pages 5937--5947, Abu Dhabi, United Arab Emirates. Association for Computational Linguistics.

\bibitem[{Dong et~al.(2023)Dong, Li, Dai, Zheng, Wu, Chang, Sun, Xu, Li, and Sui}]{dong2023survey}
Qingxiu Dong, Lei Li, Damai Dai, Ce~Zheng, Zhiyong Wu, Baobao Chang, Xu~Sun, Jingjing Xu, Lei Li, and Zhifang Sui. 2023.
\newblock \href {http://arxiv.org/abs/2301.00234} {A survey on in-context learning}.

\bibitem[{Gu et~al.(2024)Gu, Zhou, Han, Liu, Wang, and Wang}]{gu-etal-2024-pokemqa}
Hengrui Gu, Kaixiong Zhou, Xiaotian Han, Ninghao Liu, Ruobing Wang, and Xin Wang. 2024.
\newblock \href {https://doi.org/10.18653/v1/2024.acl-long.438} {{P}oke{MQA}: Programmable knowledge editing for multi-hop question answering}.
\newblock In \emph{Proceedings of the 62nd Annual Meeting of the Association for Computational Linguistics (Volume 1: Long Papers)}, pages 8069--8083, Bangkok, Thailand. Association for Computational Linguistics.

\bibitem[{Gupta et~al.(2023)Gupta, Mondal, Sheshadri, Zhao, Li, Wiegreffe, and Tandon}]{gupta2023editing}
Anshita Gupta, Debanjan Mondal, Akshay Sheshadri, Wenlong Zhao, Xiang Li, Sarah Wiegreffe, and Niket Tandon. 2023.
\newblock Editing common sense in transformers.
\newblock In \emph{Proceedings of the 2023 Conference on Empirical Methods in Natural Language Processing}, pages 8214--8232.

\bibitem[{Hartvigsen et~al.(2022)Hartvigsen, Sankaranarayanan, Palangi, Kim, and Ghassemi}]{hartvigsen2023aging}
Thomas Hartvigsen, Swami Sankaranarayanan, Hamid Palangi, Yoon Kim, and Marzyeh Ghassemi. 2022.
\newblock Aging with grace: Lifelong model editing with discrete key-value adaptors.
\newblock In \emph{NeurIPS 2022 Workshop on Robustness in Sequence Modeling}.

\bibitem[{Huang et~al.(2022)Huang, Shen, Zhang, Zhou, Rong, and Xiong}]{huang2023transformerpatcher}
Zeyu Huang, Yikang Shen, Xiaofeng Zhang, Jie Zhou, Wenge Rong, and Zhang Xiong. 2022.
\newblock Transformer-patcher: One mistake worth one neuron.
\newblock In \emph{The Eleventh International Conference on Learning Representations}.

\bibitem[{Izacard et~al.(2022)Izacard, Caron, Hosseini, Riedel, Bojanowski, Joulin, and Grave}]{izacard2022unsupervised}
Gautier Izacard, Mathilde Caron, Lucas Hosseini, Sebastian Riedel, Piotr Bojanowski, Armand Joulin, and Edouard Grave. 2022.
\newblock \href {https://openreview.net/forum?id=jKN1pXi7b0} {Unsupervised dense information retrieval with contrastive learning}.
\newblock \emph{Transactions on Machine Learning Research}.

\bibitem[{Khattab et~al.(2022)Khattab, Santhanam, Li, Hall, Liang, Potts, and Zaharia}]{khattab2022demonstrate}
Omar Khattab, Keshav Santhanam, Xiang~Lisa Li, David Hall, Percy Liang, Christopher Potts, and Matei Zaharia. 2022.
\newblock Demonstrate-search-predict: Composing retrieval and language models for knowledge-intensive nlp.
\newblock \emph{arXiv preprint arXiv:2212.14024}.

\bibitem[{Lan et~al.(2022)Lan, He, Jiang, Jiang, Zhao, and Wen}]{Lan-KBQASurvey-2021}
Yunshi Lan, Gaole He, Jinhao Jiang, Jing Jiang, Wayne~Xin Zhao, and Ji-Rong Wen. 2022.
\newblock Complex knowledge base question answering: A survey.
\newblock \emph{IEEE Transactions on Knowledge and Data Engineering}.

\bibitem[{Lei et~al.(2023)Lei, Li, Wei, He, Huang, Zhao, and Liu}]{lei-etal-2023-s3hqa}
Fangyu Lei, Xiang Li, Yifan Wei, Shizhu He, Yiming Huang, Jun Zhao, and Kang Liu. 2023.
\newblock \href {https://doi.org/10.18653/v1/2023.acl-short.147} {{S}3{HQA}: A three-stage approach for multi-hop text-table hybrid question answering}.
\newblock In \emph{Proceedings of the 61st Annual Meeting of the Association for Computational Linguistics (Volume 2: Short Papers)}, pages 1731--1740, Toronto, Canada. Association for Computational Linguistics.

\bibitem[{Li et~al.(2023{\natexlab{a}})Li, Li, Song, Yang, Ma, and Yu}]{li2023pmet}
Xiaopeng Li, Shasha Li, Shezheng Song, Jing Yang, Jun Ma, and Jie Yu. 2023{\natexlab{a}}.
\newblock \href {http://arxiv.org/abs/2308.08742} {Pmet: Precise model editing in a transformer}.

\bibitem[{Li et~al.(2023{\natexlab{b}})Li, Bubeck, Eldan, Giorno, Gunasekar, and Lee}]{li2023textbooks}
Yuanzhi Li, Sébastien Bubeck, Ronen Eldan, Allie~Del Giorno, Suriya Gunasekar, and Yin~Tat Lee. 2023{\natexlab{b}}.
\newblock \href {http://arxiv.org/abs/2309.05463} {Textbooks are all you need ii: phi-1.5 technical report}.

\bibitem[{Lin et~al.(2018)Lin, Socher, and Xiong}]{lin-etal-2018-multi}
Xi~Victoria Lin, Richard Socher, and Caiming Xiong. 2018.
\newblock \href {https://doi.org/10.18653/v1/D18-1362} {Multi-hop knowledge graph reasoning with reward shaping}.
\newblock In \emph{Proceedings of the 2018 Conference on Empirical Methods in Natural Language Processing}, pages 3243--3253, Brussels, Belgium. Association for Computational Linguistics.

\bibitem[{Mavi et~al.(2022)Mavi, Jangra, and Jatowt}]{mavi2022survey}
Vaibhav Mavi, Anubhav Jangra, and Adam Jatowt. 2022.
\newblock A survey on multi-hop question answering and generation.
\newblock \emph{arXiv preprint arXiv:2204.09140}.

\bibitem[{Meng et~al.(2022)Meng, Bau, Andonian, and Belinkov}]{NEURIPS2022_6f1d43d5}
Kevin Meng, David Bau, Alex Andonian, and Yonatan Belinkov. 2022.
\newblock \href {https://proceedings.neurips.cc/paper_files/paper/2022/file/6f1d43d5a82a37e89b0665b33bf3a182-Paper-Conference.pdf} {Locating and editing factual associations in gpt}.
\newblock In \emph{Advances in Neural Information Processing Systems}, volume~35, pages 17359--17372. Curran Associates, Inc.

\bibitem[{Meng et~al.(2023)Meng, Sen~Sharma, Andonian, Belinkov, and Bau}]{meng2022memit}
Kevin Meng, Arnab Sen~Sharma, Alex Andonian, Yonatan Belinkov, and David Bau. 2023.
\newblock Mass editing memory in a transformer.
\newblock \emph{The Eleventh International Conference on Learning Representations (ICLR)}.

\bibitem[{Mitchell et~al.(2022{\natexlab{a}})Mitchell, Lin, Bosselut, Finn, and Manning}]{mitchell2022fast}
Eric Mitchell, Charles Lin, Antoine Bosselut, Chelsea Finn, and Christopher~D Manning. 2022{\natexlab{a}}.
\newblock \href {https://openreview.net/forum?id=0DcZxeWfOPt} {Fast model editing at scale}.
\newblock In \emph{International Conference on Learning Representations}.

\bibitem[{Mitchell et~al.(2022{\natexlab{b}})Mitchell, Lin, Bosselut, Finn, and Manning}]{mitchell2022memory}
Eric Mitchell, Charles Lin, Antoine Bosselut, Chelsea Finn, and Christopher~D. Manning. 2022{\natexlab{b}}.
\newblock \href {https://arxiv.org/pdf/2206.06520.pdf} {Memory-based model editing at scale}.
\newblock In \emph{International Conference on Machine Learning}.

\bibitem[{Nie et~al.(2024)Nie, Zhang, Wang, and Liu}]{nie2024code}
Zhijie Nie, Richong Zhang, Zhongyuan Wang, and Xudong Liu. 2024.
\newblock Code-style in-context learning for knowledge-based question answering.
\newblock In \emph{Proceedings of the AAAI Conference on Artificial Intelligence}, volume~38, pages 18833--18841.

\bibitem[{Pan et~al.(2023)Pan, Luo, Wang, Chen, Wang, and Wu}]{pan2023unifying}
Shirui Pan, Linhao Luo, Yufei Wang, Chen Chen, Jiapu Wang, and Xindong Wu. 2023.
\newblock Unifying large language models and knowledge graphs: A roadmap.
\newblock \emph{arXiv preprint arXiv:2306.08302}.

\bibitem[{Piscopo and Simperl(2019)}]{10.1145/3306446.3340822}
Alessandro Piscopo and Elena Simperl. 2019.
\newblock \href {https://doi.org/10.1145/3306446.3340822} {What we talk about when we talk about wikidata quality: a literature survey}.
\newblock In \emph{Proceedings of the 15th International Symposium on Open Collaboration}, OpenSym '19, New York, NY, USA. Association for Computing Machinery.

\bibitem[{Press et~al.(2023)Press, Zhang, Min, Schmidt, Smith, and Lewis}]{press-etal-2023-measuring}
Ofir Press, Muru Zhang, Sewon Min, Ludwig Schmidt, Noah Smith, and Mike Lewis. 2023.
\newblock \href {https://doi.org/10.18653/v1/2023.findings-emnlp.378} {Measuring and narrowing the compositionality gap in language models}.
\newblock In \emph{Findings of the Association for Computational Linguistics: EMNLP 2023}, pages 5687--5711, Singapore. Association for Computational Linguistics.

\bibitem[{Rae et~al.(2022)Rae, Borgeaud, Cai, Millican, Hoffmann, Song, Aslanides, Henderson, Ring, Young, Rutherford, Hennigan, Menick, Cassirer, Powell, van~den Driessche, Hendricks, Rauh, Huang, Glaese, Welbl, Dathathri, Huang, Uesato, Mellor, Higgins, Creswell, McAleese, Wu, Elsen, Jayakumar, Buchatskaya, Budden, Sutherland, Simonyan, Paganini, Sifre, Martens, Li, Kuncoro, Nematzadeh, Gribovskaya, Donato, Lazaridou, Mensch, Lespiau, Tsimpoukelli, Grigorev, Fritz, Sottiaux, Pajarskas, Pohlen, Gong, Toyama, de~Masson~d'Autume, Li, Terzi, Mikulik, Babuschkin, Clark, de~Las~Casas, Guy, Jones, Bradbury, Johnson, Hechtman, Weidinger, Gabriel, Isaac, Lockhart, Osindero, Rimell, Dyer, Vinyals, Ayoub, Stanway, Bennett, Hassabis, Kavukcuoglu, and Irving}]{rae2022scaling}
Jack~W. Rae, Sebastian Borgeaud, Trevor Cai, Katie Millican, Jordan Hoffmann, Francis Song, John Aslanides, Sarah Henderson, Roman Ring, Susannah Young, Eliza Rutherford, Tom Hennigan, Jacob Menick, Albin Cassirer, Richard Powell, George van~den Driessche, Lisa~Anne Hendricks, Maribeth Rauh, Po-Sen Huang, Amelia Glaese, Johannes Welbl, Sumanth Dathathri, Saffron Huang, Jonathan Uesato, John Mellor, Irina Higgins, Antonia Creswell, Nat McAleese, Amy Wu, Erich Elsen, Siddhant Jayakumar, Elena Buchatskaya, David Budden, Esme Sutherland, Karen Simonyan, Michela Paganini, Laurent Sifre, Lena Martens, Xiang~Lorraine Li, Adhiguna Kuncoro, Aida Nematzadeh, Elena Gribovskaya, Domenic Donato, Angeliki Lazaridou, Arthur Mensch, Jean-Baptiste Lespiau, Maria Tsimpoukelli, Nikolai Grigorev, Doug Fritz, Thibault Sottiaux, Mantas Pajarskas, Toby Pohlen, Zhitao Gong, Daniel Toyama, Cyprien de~Masson~d'Autume, Yujia Li, Tayfun Terzi, Vladimir Mikulik, Igor Babuschkin, Aidan Clark, Diego de~Las~Casas, Aurelia Guy, Chris Jones,
  James Bradbury, Matthew Johnson, Blake Hechtman, Laura Weidinger, Iason Gabriel, William Isaac, Ed~Lockhart, Simon Osindero, Laura Rimell, Chris Dyer, Oriol Vinyals, Kareem Ayoub, Jeff Stanway, Lorrayne Bennett, Demis Hassabis, Koray Kavukcuoglu, and Geoffrey Irving. 2022.
\newblock \href {http://arxiv.org/abs/2112.11446} {Scaling language models: Methods, analysis \& insights from training gopher}.

\bibitem[{Rawat et~al.(2020)Rawat, Zhu, Li, Yu, Zaheer, Kumar, and Bhojanapalli}]{52308}
Ankit~Singh Rawat, Chen Zhu, Daliang Li, Felix Yu, Manzil Zaheer, Sanjiv Kumar, and Srinadh Bhojanapalli. 2020.
\newblock Modifying memories in transformer models.
\newblock In \emph{International Conference on Machine Learning (ICML) 2021}.

\bibitem[{Rawte et~al.(2023)Rawte, Sheth, and Das}]{rawte2023survey}
Vipula Rawte, Amit Sheth, and Amitava Das. 2023.
\newblock A survey of hallucination in large foundation models.
\newblock \emph{arXiv preprint arXiv:2309.05922}.

\bibitem[{Sen et~al.(2023)Sen, Mavadia, and Saffari}]{sen-etal-2023-knowledge}
Priyanka Sen, Sandeep Mavadia, and Amir Saffari. 2023.
\newblock \href {https://doi.org/10.18653/v1/2023.nlrse-1.1} {Knowledge graph-augmented language models for complex question answering}.
\newblock In \emph{Proceedings of the 1st Workshop on Natural Language Reasoning and Structured Explanations (NLRSE)}, pages 1--8, Toronto, Canada. Association for Computational Linguistics.

\bibitem[{Sinitsin et~al.(2020)Sinitsin, Plokhotnyuk, Pyrkin, Popov, and Babenko}]{Sinitsin2020Editable}
Anton Sinitsin, Vsevolod Plokhotnyuk, Dmitry Pyrkin, Sergei Popov, and Artem Babenko. 2020.
\newblock \href {https://openreview.net/forum?id=HJedXaEtvS} {Editable neural networks}.
\newblock In \emph{International Conference on Learning Representations}.

\bibitem[{Vrande{\v{c}}i{\'c} and Kr{\"o}tzsch(2014)}]{vrandevcic2014wikidata}
Denny Vrande{\v{c}}i{\'c} and Markus Kr{\"o}tzsch. 2014.
\newblock Wikidata: a free collaborative knowledgebase.
\newblock \emph{Communications of the ACM}, 57(10):78--85.

\bibitem[{Wang and Komatsuzaki(2021)}]{gpt-j}
Ben Wang and Aran Komatsuzaki. 2021.
\newblock {GPT-J-6B: A 6 Billion Parameter Autoregressive Language Model}.
\newblock \url{https://github.com/kingoflolz/mesh-transformer-jax}.

\bibitem[{Wei et~al.(2022)Wei, Wang, Schuurmans, Bosma, ichter, Xia, Chi, Le, and Zhou}]{NEURIPS2022_9d560961}
Jason Wei, Xuezhi Wang, Dale Schuurmans, Maarten Bosma, brian ichter, Fei Xia, Ed~Chi, Quoc~V Le, and Denny Zhou. 2022.
\newblock \href {https://proceedings.neurips.cc/paper_files/paper/2022/file/9d5609613524ecf4f15af0f7b31abca4-Paper-Conference.pdf} {Chain-of-thought prompting elicits reasoning in large language models}.
\newblock In \emph{Advances in Neural Information Processing Systems}, volume~35, pages 24824--24837. Curran Associates, Inc.

\bibitem[{Welbl et~al.(2018)Welbl, Stenetorp, and Riedel}]{welbl-etal-2018-constructing}
Johannes Welbl, Pontus Stenetorp, and Sebastian Riedel. 2018.
\newblock \href {https://doi.org/10.1162/tacl_a_00021} {Constructing datasets for multi-hop reading comprehension across documents}.
\newblock \emph{Transactions of the Association for Computational Linguistics}, 6:287--302.

\bibitem[{Yang et~al.(2018)Yang, Qi, Zhang, Bengio, Cohen, Salakhutdinov, and Manning}]{yang-etal-2018-hotpotqa}
Zhilin Yang, Peng Qi, Saizheng Zhang, Yoshua Bengio, William Cohen, Ruslan Salakhutdinov, and Christopher~D. Manning. 2018.
\newblock \href {https://doi.org/10.18653/v1/D18-1259} {{H}otpot{QA}: A dataset for diverse, explainable multi-hop question answering}.
\newblock In \emph{Proceedings of the 2018 Conference on Empirical Methods in Natural Language Processing}, pages 2369--2380, Brussels, Belgium. Association for Computational Linguistics.

\bibitem[{Yao et~al.(2023)Yao, Wang, Tian, Cheng, Li, Deng, Chen, and Zhang}]{DBLP:journals/corr/abs-2305-13172}
Yunzhi Yao, Peng Wang, Bozhong Tian, Siyuan Cheng, Zhoubo Li, Shumin Deng, Huajun Chen, and Ningyu Zhang. 2023.
\newblock \href {https://aclanthology.org/2023.emnlp-main.632} {Editing large language models: Problems, methods, and opportunities}.
\newblock In \emph{Proceedings of the 2023 Conference on Empirical Methods in Natural Language Processing}, pages 10222--10240, Singapore. Association for Computational Linguistics.

\bibitem[{Yin et~al.(2016)Yin, Lu, Li, and Ben}]{yin-etal-2016-neural}
Pengcheng Yin, Zhengdong Lu, Hang Li, and Kao Ben. 2016.
\newblock \href {https://doi.org/10.18653/v1/W16-0105} {Neural enquirer: Learning to query tables in natural language}.
\newblock In \emph{Proceedings of the Workshop on Human-Computer Question Answering}, pages 29--35, San Diego, California. Association for Computational Linguistics.

\bibitem[{Zhang et~al.(2024)Zhang, Ye, Liu, Ren, Wu, and Chen}]{DBLP:journals/corr/abs-2402-13593}
Mengqi Zhang, Xiaotian Ye, Qiang Liu, Pengjie Ren, Shu Wu, and Zhumin Chen. 2024.
\newblock \href {https://doi.org/10.48550/ARXIV.2402.13593} {Knowledge graph enhanced large language model editing}.
\newblock \emph{CoRR}, abs/2402.13593.

\bibitem[{Zhong et~al.(2023)Zhong, Wu, Manning, Potts, and Chen}]{zhong2023mquake}
Zexuan Zhong, Zhengxuan Wu, Christopher Manning, Christopher Potts, and Danqi Chen. 2023.
\newblock \href {https://aclanthology.org/2023.emnlp-main.971} {{MQ}u{AKE}: Assessing knowledge editing in language models via multi-hop questions}.
\newblock In \emph{Proceedings of the 2023 Conference on Empirical Methods in Natural Language Processing}, pages 15686--15702, Singapore. Association for Computational Linguistics.

\end{thebibliography}

\appendix
\newpage
\section{Implementation}
\subsection{Dataset Statistics} \label{A1}
Table \ref{data_statistics} provides a summary of the statistics for the MQuAKE-CF and MQuAKE-T datasets. 
% Table generated by Excel2LaTeX from sheet 'mquake_statistics'
\begin{table}[h]
  \centering
  \setlength{\tabcolsep}{2pt}
    \begin{tabular}{llrrrr}
    \toprule
          & \#Edits & \multicolumn{1}{l}{2-hop} & \multicolumn{1}{l}{3-hop} & \multicolumn{1}{l}{4-hop} & \multicolumn{1}{l}{Total} \\
    \midrule
    \multirow{5}[0]{*}{MQuaKE-CF} & \multicolumn{1}{l}{1} & 513   & 356   & 224   & 1,093 \\
          & \multicolumn{1}{l}{2} & 487   & 334   & 246   & 1,067 \\
          & \multicolumn{1}{l}{3} &   -    & 310   & 262   & 572 \\
          & \multicolumn{1}{l}{4} &   -    &   -    & 268   & 268 \\
          & All   & 1,000 & 1,000 & 1,000 & 3,000 \\
    \midrule
    MQuaKE-T & 1 (All) & 1,421 & 445   & 2     & 1,868 \\
    \bottomrule
    \end{tabular}%
    \caption{Statistics of MQuAKE dataset \cite{zhong2023mquake}.}
  \label{data_statistics}%
\end{table}%

\begin{table}[t]
    \centering
  \setlength{\tabcolsep}{6pt}
\begin{tabular}{@{}lcccc@{}}
\toprule
        & k=100         & k=1000       & k=3000       & Average        \\ \midrule
Top-4   & 15.6          & \textbf{9.1} & 7.2          & 10.63          \\
Top-5   & \textbf{16.8} & 8.3          & 6.9          & 10.67          \\
Top-6   & 16.6          & 8.5          & 7.4          & \textbf{10.83} \\
Top-10  & 15.3          & 9.0          & \textbf{8.0} & 10.77          \\
Top-100 & 8.2           & 4.7          & 3.7          & 5.53           \\ \bottomrule
\end{tabular}%
\caption{Hyperparameter search for top-$x$ in Vicuna-based QA systems on the MQuAKE-CF dataset.}
\label{topk_hyper}
\end{table}

\subsection{Hyperparmers Settings}
To ensure reproducibility, we set the temperature to zero in all experiments. Table \ref{topk_hyper} shows that retrieving the top-6 edited facts from external memory provides the best average performance on the MQuAKE-CF dataset\footnote{Tested only on the first question of each test instance, rather than all three} for $k>1$. Consequently, we include top-6 edited facts in the prompt for subsequent experiments on this dataset when $k>1$. Similarly, for the MQuAKE-T dataset when $k>1$, we opted to incorporate the top-1 edited fact in the prompt.
\subsection{Predefined Relations Utilized in the Prompts for Relation Chain and Fact Triple Extraction}
After filtering by GPT-3.5-Turbo, the first 50 relations utilized in MQuAKE-CF dataset are:
['country of origin','sport', 'country of citizenship', 'capital', 'continent', 'official language', 'head of state', 'head of government', 'creator', 'country', 'author', 'headquarters location', 'place of birth','spouse', 'director / manager','religion or worldview', 'genre', 'work location', 'performer','manufacturer', 'developer', 'place of death', 'employer', 'educated at','member of sports team', 'head coach', 'languages spoken, written or signed', 'notable work', 'child', 'founded by', 'location', 'chief executive officer', 'original broadcaster', 'chairperson', 'occupation', 'position played on team / speciality','member of', 'language of work or name', 'director', 'league', 'home venue', 'native language', 'composer', 'place of origin (Switzerland)', 'officeholder','religious order', 'publisher', 'original language of film or TV show', 'ethnic group','military branch'].

After GPT-3.5-Turbo filtering, the MQuAKE-T dataset includes a total of 35 relations. The relation list is 
['head of government', 'country of citizenship', 'head of state', 'country of origin', 'country', 'headquarters location', 'location', 'sport', 'performer', 'genre', 'developer', 'employer', 'manufacturer', 'place of death', 'place of birth', 'author', 'member of', 'capital', 'member of sports team', 'chief executive officer', 'notable work', 'director / manager', 'original broadcaster', 'creator', 'work location', 'educated at', 'located in the administrative territorial entity', 'head coach', 'place of publication', 'location of formation', 'director', 'producer', 'transport network', 'continent', 'child']

% \subsubsection{Implementation of LLM-Based QA} \label{append:llm-qa}

% We also implement the in-context learning \cite{dong2023survey} for LLM-based QA. We provide 4 samples in the prompt for MQuAKE-CF: one 1-edit sample, one 2-edit sample, one 3-edit sample, and one 4-edit sample. When k>=100, we retrieve 4 relevant edit facts for each test sample. When k=1, the prompt consists of all the relevant facts for a specific test sample, given that the edit facts in the memory is less than 5. %Regarding MQuAKE-T, we incorporate 3 samples into in the prompt.
% %We include 3 samples in the prompt for MQuAKE-T. 

\subsection{Prompt Setup and Post-Processing} \label{A3}

\begin{figure}[h]
    \centering
    \includegraphics[width=\columnwidth]{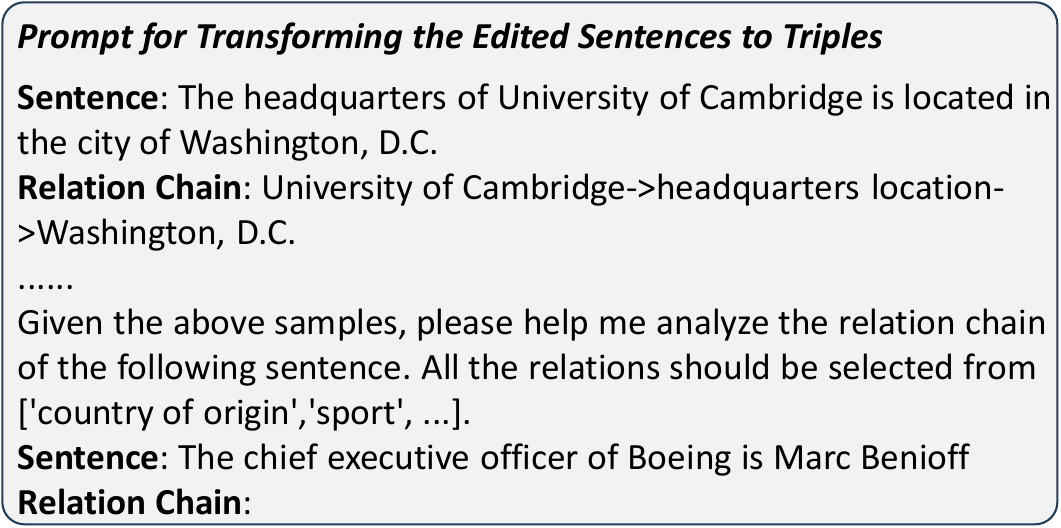}
    \caption{The prompt used for transforming edited fact sentences to triples.}
    \label{triple_prompt}
\end{figure}

\begin{figure}[h]
    \centering
    \includegraphics[width=\columnwidth]{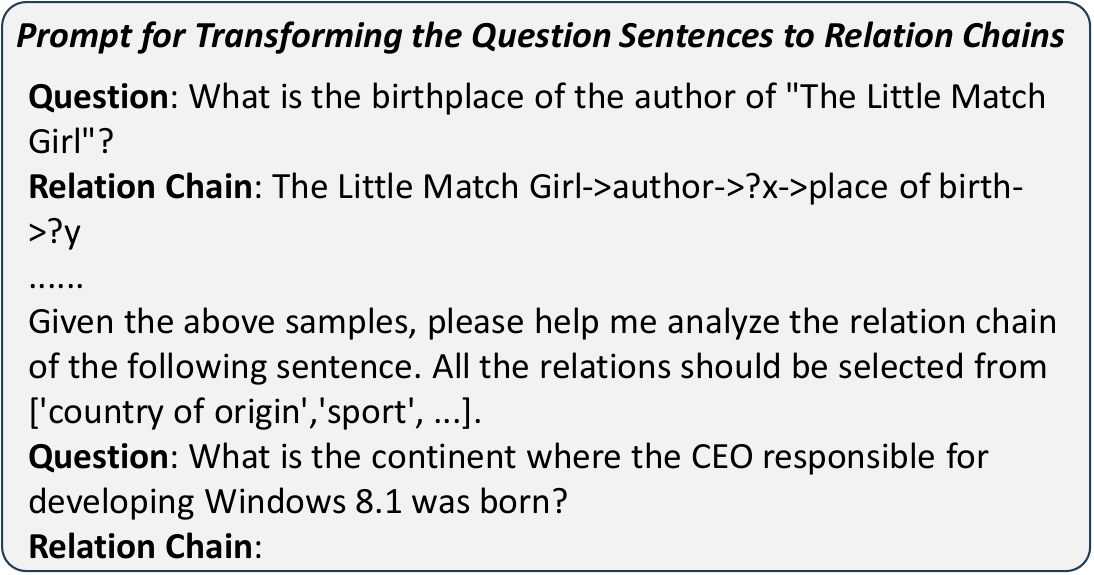}
    \caption{The prompt used for transforming question sentences to relation chains.}
    \label{relation_chain_prompt}
\end{figure}

\begin{figure}[h]
    \centering
    \includegraphics[width=\columnwidth]{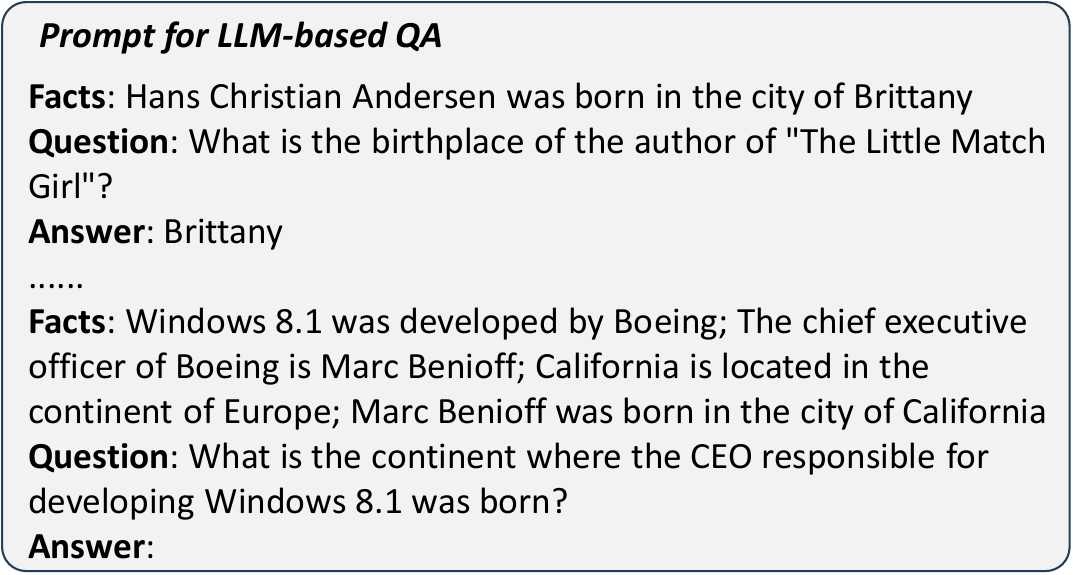}
    \caption{The prompt used in LLM-based QA.}
    \label{llm_qa_prompt}
\end{figure}

The prompts used for edited fact triple extraction, relation chain extraction, and LLM-based QA are depicted in Figures~\ref{triple_prompt}, \ref{relation_chain_prompt}, and \ref{llm_qa_prompt}. The edited triple can be regarded as a specialized relation chain, with only one relation between entities and all entities known. All samples in the prompt are selected from the complete MQuAKE-CF dataset, ensuring they are distinct from the test samples.

To improve the performance of LLMs in extracting relation chains and ensure that outputs conform to a specified format, we employ a 4-shot learning approach for the MQuAKE-CF dataset and a 3-shot learning approach for the MQuAKE-T dataset. For MQuAKE-CF, the approach involves presenting the model with samples of one 2-hop question, one 3-hop question, and two 4-hop questions. For MQuAKE-T, the model is presented with one 2-hop question, one 3-hop question, and one 4-hop question. 

To address the limitations of GPT-J and Vicuna in conforming to the desired output format, we establish a heuristic rule for extracting essential information from their outputs. For instance, in the context of relation chain extraction, this heuristic is outlined as follows:
\begin{itemize}
    \item Narrow the attention to the output sentence containing the "->" indicator.
    \item Divide the sentence based on the "->" delimiter.
    \item Regard the initial segment as the predicted entity. Subsequently, process the following segments sequentially as relations, provided they do not begin with "?".
\end{itemize}

\subsection{Strategies for Managing Sequence Errors in Extracting Fact Triples}
While LLMs consistently identifies relations accurately—such as 'head of state,' 'chief of department,' and 'head of government'—it often makes errors in their sequencing. To address this, we employ Spacy\footnote{\url{https://spacy.io/}} to detect instances where the object of an edited triple is not a person. If it is not, we adjust the sequence of the object and subject in the triple accordingly.

\section{The Distinctions Between Our GMeLLo and Other Methods}\label{dist}
While both GMeLLo and MeLLo \cite{zhong2023mquake} are memory-based models targeting multi-hop question answering in an evolving environment, they differ in the following aspects:
\begin{itemize}
    \item MeLLo employs in-context learning to direct LLMs in splitting the question into sub-questions, answering each, and verifying against relevant edited facts for contradictions. In contrast, GMeLLo retrieves pertinent edited facts for the multi-hop question and presents them alongside the question to LLMs for answering.
    \item Except storing edited facts as isolated sentences in an external memory, we leverage LLMs to translate these sentences into triples and update the KG. In addition to obtaining an answer from LLMs, we utilize KBQA to enhance the precision of multi-hop question answering within an evolving environment.
\end{itemize}

Recently, the advent of LLMs has spurred the development of LLM-based KBQA systems \cite{baek-etal-2023-knowledge-augmented, sen-etal-2023-knowledge, nie2024code}. However, our GMeLLo are different from these works in the following aspects:
\begin{itemize}
    \item Firstly, we consider question answering in a dynamic environment, where changes in the knowledge graph need to accounted for, whereas they do not.
    \item Secondly, we focus on multi-hop questions, whereas they deal with standard KBQA tasks, including intersection and difference questions etc.
    \item Thirdly, the KBQA and LLM-based QA are handled separately, using the KBQA answer as the final answer. In contrast, they retrieve triples from the knowledge graph and incorporate them into the prompt to guide LLM-based QA.
\end{itemize}
\section{Multi-Hop Performance Analysis}

We study the breakdown of performance on the MQuAKE-CF dataset with respect to the number of hops when $k=1$. Table \ref{performance_by_hops} provides the hop-specific performance of different methods. Although MQuAKE did not provide the hop performance for MeLLo, it can be inferred that the average hop performance should not exceed 65.6\%, given that the overall performance is 20.3\%.

\begin{table}[t]
  \centering
  \begin{tabular}{llcccc}
    \toprule
    \multicolumn{1}{c}{\multirow{2}[0]{*}{Model}} & \multicolumn{1}{c}{\multirow{2}[0]{*}{Method}} & \multicolumn{4}{c}{Number of Hops} \\
    \cmidrule(lr){3-6}
          &       & \multicolumn{1}{l}{2} & \multicolumn{1}{l}{3} & \multicolumn{1}{l}{4} & \multicolumn{1}{l}{Avg} \\
    \midrule
    \multirow{4}{*}{GPT-J-6B} & MEND  & 13.9 & 11.3 & 9.5 & 11.5 \\
      & MEMIT  & 22.5 & 6.0 & 8.4 & 12.3 \\
      & MeLLo  & - & - & - & 20.3 \\
      & GMeLLo & 89.5 & 73.7 & 65.6 & 76.3 \\
    \bottomrule
  \end{tabular}%
  \caption{The breakdown performance on the MQuAKE-CF dataset with respect to the number of hops when $k=1$.}
  \label{performance_by_hops}
\end{table}
\end{document}